\documentclass[runningheads,a4paper]{llncs}
\usepackage[utf8]{inputenc}

\usepackage{cite}

\usepackage{graphicx}

\usepackage[cmex10]{amsmath}

\usepackage{array}
\usepackage{arydshln}

\usepackage{eqparbox}

\usepackage{amsfonts}
\usepackage{amssymb}
\usepackage{balance}
\usepackage{enumerate}
\usepackage{latexsym}
\pdfoutput=1
\usepackage{hyperref}
\hypersetup{
    colorlinks=true,
    linkcolor=black,
    citecolor=black,
    filecolor=black,
    urlcolor=black,
}

\usepackage{floatrow}

\usepackage[cmex10]{amsmath}
\usepackage{mathrsfs}
\usepackage{multirow}

\usepackage{leftidx}
\usepackage{stmaryrd}
\usepackage{dsfont}

\usepackage{algorithm}
\usepackage{algorithmic}

\usepackage{hhline}
\usepackage{color}
\newcommand{\va}{\mathbf{a}}
\newcommand{\vb}{\mathbf{b}}

\newcommand{\vm}{\mathbf{m}}

\newcommand{\matr}[1]{\mathbf{#1}}
\newcommand{\mA}{\matr{A}}
\newcommand{\mB}{\matr{B}}
\newcommand{\mC}{\matr{C}}

\newcommand{\mV}{\matr{V}}

\newcommand{\mM}{\matr{M}}
\newcommand{\mD}{\matr{D}}

\newcommand{\T}{{\sf T}}        

\newcommand{\diag}[1]{\text{Diag}\left(#1\right)}   

\makeatletter
\newcommand{\argmin}{\mathop{\operator@font argmin}}

\newcommand{\tens}[1]{\boldsymbol{\mathcal{#1}}}

\newcommand{\tM}{\tens{M}}

\definecolor{gris}{gray}{0.90}
\definecolor{gris25}{gray}{0.90}
\definecolor{americanrose}{rgb}{1.0, 0.01, 0.24}
\definecolor{bostonuniversityred}{rgb}{0.8, 0.0, 0.0}
\definecolor{shamrockgreen}{rgb}{0.0, 0.62, 0.38}
\definecolor{selectiveyellow}{rgb}{1.0, 0.73, 0.0}
\definecolor{royalblue}{rgb}{0.25, 0.41, 0.88}
\definecolor{ashgrey}{rgb}{0.7, 0.75, 0.71}
\definecolor{burgundy}{RGB}{159,29,53}
\definecolor{darkgreen}{RGB}{18,53,26}
\definecolor{lightblue}{RGB}{102,217,255}
\definecolor{fakeorange}{RGB}{255,140,102}
\definecolor{arylideyellow}{rgb}{0.91, 0.84, 0.42}
\definecolor{bananayellow}{rgb}{1.0, 0.88, 0.21}
\definecolor{gris_f}{gray}{0.35}
\definecolor{bordure}{rgb}{0.09,0.17,0.68}
\definecolor{aquamarine}{rgb}{0.5, 1.0, 0.83}
\definecolor{apricot}{rgb}{0.98, 0.81, 0.69}
\definecolor{babyblue}{rgb}{0.54, 0.81, 0.94}
\definecolor{uipoppy}{RGB}{225, 64, 5}
\definecolor{uipaleblue}{RGB}{96,123,139}
\definecolor{uiblack}{RGB}{0, 0, 0}
\definecolor{decoda}{RGB}{0,153, 0}
\definecolor{lightgreen}{rgb}{0.56, 0.93, 0.56}
\definecolor{blue_f}{rgb}{0.2, 0.2, 0.6}

\usepackage{color}
\definecolor{darkred}{rgb}{0.7,0,0}
\definecolor{darkgreen}{rgb}{0,0.46,0}

\makeatletter
\renewcommand{\maketag@@@}[1]{\hbox{\m@th\normalsize\normalfont#1}}%
\makeatother

\urldef{\mailsa}\path|{jeremy.cohen}@umons.ac.be|    
\newcommand{\keywords}[1]{\par\addvspace\baselineskip
\noindent\keywordname\enspace\ignorespaces#1}
\toctitle{Curve Registered Coupled Low Rank Factorization}

\begin{document}
%
\mainmatter

\title{Curve Registered Coupled Low Rank Factorization}

\titlerunning{Curve Registered Coupled Low Rank Factorization}

\author{Jeremy~Emile~Cohen\inst{1}
\and Rodrigo~Cabral~Farias\inst{2}
\and Bertrand~Rivet\inst{3}~
}

\institute{Dept. of Mathematics and Operational Research, \\ 
Rue de Houdain 9, Facult\'e polytechnique, Universit\'e de Mons 
\\ \mailsa 
\and Univ. C\^ote d'Azur, CNRS, I3S, 06900 Sophia-Antipolis
\and Univ. Grenoble Alpes, CNRS, Grenoble INP\footnote{Institute of Engineering Univ. Grenoble Alpes}, GIPSA-lab, 38000 Grenoble, France}


\maketitle

\begin{abstract}
We propose an extension of the canonical polyadic (CP) tensor model where one of the latent factors is allowed to vary through data slices in a constrained way. The components of the latent factors, which we want to retrieve from data, can vary from one slice to another up to a diffeomorphism. We suppose that the diffeomorphisms are also unknown, thus merging curve registration and tensor decomposition in one model, which we call registered CP. We present an algorithm to retrieve both the latent factors and the diffeomorphism, which is assumed to be in a parametrized form. At the end of the paper, we show simulation results comparing registered CP with other models from the literature.
\end{abstract}

\keywords{Tensor decompositions, Curve registration, Data fusion.} 

%

\section{Introduction}\label{intro-sec}

Joint decomposition models such as the canonical polyadic (CP) tensor decomposition~\cite{comon2009tensor} allow to blindly extract patterns of underlying hidden phenomena from a block of data measurements based on their algebraic properties without statistical assumptions.  
Thanks to their uniqueness properties under mild conditions~\cite{comon2009tensor}, tensor decompositions have been applied in many domains: neurosciences \cite{BeckACGWM15:spmag}, chemometrics~\cite{smilde2005multi} and  digital communications~\cite{sidiropoulos2000blind} to name a few.

To retrieve the latent patterns without statistical assumptions, the number of free parameters must be rather low (\textit{i.e.} the number of latent patterns is small with respect to the data dimensions). 
For example, in the CP model for a 3-way data block, $\tM\in\mathbb{R}^3$, each slice $ \mM_k $ in one of the dimensions is approximated by a rank $R$ matrix decomposition:  $ \mM_k=\mA\diag{\mC(k,:)}\mB^T  $, where $\mA=[\va_1,\dots,\va_R]$, $\mB=[\vb_1,\dots,\vb_R]$, and $ \diag{\mC(k,:)} $ is the diagonal matrix formed with the $ k $-th row of $\mC=[\mathbf{c}_1,\dots,\mathbf{c}_R] $. 
Here the columns of these matrices are the latent patterns that we are searching for and the fundamental constraint is that the matrix factors $ \mA $ and $ \mB $ are exactly the same as $ k $ varies. 
Clearly, the model for the slices in any of the 3-ways of the CP decomposition corresponds to a coupled matrix decomposition where the $\mA$ and $\mB$ matrix factors are shared. 
Other models with less stringent coupling constraints have been considered in the literature, for example, PARAFAC2~\cite{harshman1972parafac2}, Shift-PARAFAC~\cite{morup2008shift,harshman2003shifted}, soft non-negative matrix co-factorization~\cite{seichepine2014soft} or probabilistic couplings~\cite{CabrCC16:tsp}.

In this paper, we are also interested in such a less constrained decomposition, where one of the matrix factors, $ \mB$ for example, is allowed to change over the experimental parameter $k$: $\mM_k=\mA\diag{\mC(k,:)}\mB_k^T$. 
The components of the factor from one slice to another are all similar up to a diffeomorphism, that is up to local compression and dilations. 
This can be useful, for example, in ocular-artifact removal in EEG~\cite{rivet2015multimodal} where the coupled latent signals are related to different eye blinks or saccades, or in chromatography \cite{bro1999parafac2} where the latent components are time elution responses of chemical compounds on different chromatographic experiments. 
In both exemples, the patterns feature domain variations, that may differ at any given time but are similar after alignment through delay, local dilations and compressions. 

Finding the diffeomorphisms, that is, the transformations of the arguments (time or space) of the latent curves, leading to an alignment is known in statistics as curve registration~\cite{ramsay1998curve} and in signal processing as time warping~\cite{sakoe1978dynamic}. 
In curve registration one may be interested in computing the structured average~\cite{kneip1992statistical}, \textit{i.e.} an aligned mean curve, which serves as a template for trend analysis. 
In this paper, we are facing a different problem than in curve registration since the curves themselves are unknown latent functions.
By merging both curve registration and CP decompositions, we expect that the factors obtained from the joint decomposition of each slice will be retrieved with an increased accuracy when compared with other methods which do not include fully the diffeomorphism coupling information, as in Shift-PARAFAC \cite{harshman2003shifted} and PARAFAC2~\cite{morup2008shift,marini2013scream}.

In this work we propose to modify the well-known alternating least squares (ALS) algorithm for CP decomposition~\cite{comon2009tensor} to include a curve registration step on the factor containing domain variation. 
Closely related to our work, warped factor analysis (WFA) has been proposed in~\cite{hong2009warped} where curve registration is explicitly carried out using a piecewise linear model for the diffeomorphism. 
In WFA, the template curve (\textit{i.e.} the structural average which is used as reference) is contained directly in the data, which is a fundamental difference with the proposed approach. 
In our work we extend WFA (i) to a generalized diffeomorphism model, and (ii) to have a less arbitrary template curve estimated from all latent patterns by searching for a structural average curve.
To retrieve this structural average curve and the optimal diffeomorphisms, we follow an alternating approach similar to~\cite{srivastava2011registration}. 

\paragraph{Notation:}
Vectors are denoted in bold symbols $\va$, matrices as bold capital symbols $\mM$.
The $(i,j)$-th entry of matrix $\mM$ is denoted $\mM(i,j)$, its $i$-th column~$\mM(:,i)$ or $\vm_i$ and the $i$-th row  $\mM(i,:)$.
The transposition operator is denoted as~$\mM^T$. $\circ$~is the composition operator: $(f\circ g)(\cdot) = f\bigl(g(\cdot)\bigr)$.

\section{Curve registered decomposition models}\label{sec:diffeo} 
In this section we present the curve registered decomposition model through a Bayesian estimation perspective. 
We present it in three steps: Section~\ref{Sec:1} develops the measurement model and its corresponding likelihood.
Section~\ref{Sec:2} presents the registered CP derived from the maximum \textit{a posteriori} (MAP) estimator of all unknown parameters (\textit{i.e.} both measurement and coupling models).
Finally Section~\ref{parametric_warp} introduces a parametric model for the diffeomorphisms.

\subsection{Measurement model: low-rank matrix decomposition model}
\label{Sec:1}
Without loss of generality, we consider the data block to be a 3-way array,
$\tM\in\mathbb{R}^{I\times J \times K}$, such that $ K $ 2-way measurement
arrays ($\mM_k \in \mathbb{R}^{I\,\times\,J} $) of size $ I\, \times\, J $ are
available. Moreover, we suppose that each matrix is given by a rank-$ R $ factorization plus a measurement noise term:
\begin{equation}
\mM_k =\sum\limits_{r=1}^{R}{c_{k,r} \va_r \vb_r^T} +\mV_k = \mA\diag{\matr{C}(k,:)}\mB_k^T + \mV_k
\label{mat_decomp}
\end{equation}
\noindent where the rank $ R $ is supposed to be known and much smaller than the dimensions $ I $, $ J $ and $ K $. 
The factor matrices $ \mA $, $\bigl\{\mB_k\bigr\}_{1\leq k \leq K}$ and $ \mC $
are the unknown latent patterns to be retrieved and $ \mV_k $ are noise
matrices assumed to be independent from one another and with independent
elements. Note that the factor matrix $\mA$ is shared across data slices
$\mM_k$.
The elements~$v_{ijk}=\mV_k(i,j)$ of the noise matrices are assumed to be independent zero-mean normally distributed with a variance $\sigma_k^2$: $p(v_{ijk})\propto\exp\left\lbrace  -v_{ijk}^2/2\sigma_k^2 \right\rbrace$.
Without further knowledge on a relationship relating factors $\mB_k$, a natural way to retrieve the latent factors is through maximum likelihood estimation. 
This corresponds to the minimization of the cost function $ \mathcal{L} $ w.r.t. the factor matrices:

\begin{equation} 
\mathcal{L}=\sum\limits_{k=1}^{K} \frac{1}{\sigma_k^{2}}\| \mM_k-
\mA\diag{\mC(k,:)}\mB_k^T \|_\text{F}^{2}, \label{mle} 
\end{equation} 
where $ \|\cdot \|_\text{F} $ stands for the Frobenius norm.  
Minimizing~\eqref{mle} actually corresponds to computing a low rank matrix factorization of the stacked matrices $\mM_{1:K} = [\frac{1}{\sigma_n^1}\mM_1,\dots,\frac{1}{\sigma_n^K}\mM_K]$.
Therefore, there is no guarantee that the retrieved patterns will be physically interpretable, since the model is not uniquely identifiable due to rotational ambiguity.

\subsection{Registered CP from MAP formulation}
\label{Sec:2}
In what follows, factors $\mB_k$ are supposed to be similar in shape but with variations on their domain.
For example, consider that factors $ \mB_k  $ relate to time and that they are sampled versions of continuous-time signals: $\mB_k(r,j) = b_{k,r}(t_j)$.
We assume that the sampling grid points $ t_j $, with $ j\in\left\lbrace 1,\,\cdots,\,J\right\rbrace  $, are the same for all measurement matrices and we consider a normalized time period so that $ t_j\in \left[0,\,1 \right]   $. 
For any of the $K$ underlying continuous signals, domain variation can be expressed as
\begin{equation}
\forall (r,k)\in \llbracket1,R\rrbracket\times\llbracket1,K\rrbracket,\quad
b_{r,k}(t) =b^*_r\bigl(\gamma_{r,k}(t)\bigr) + w_{r,k}(t),
\label{diff_constraint}
\end{equation}
where the  functions representing the variation $ \gamma_{r,k}(t) $ are diffeomorphisms from $ \left[0,1 \right]  $ to $ \left[0,1 \right] $. 
They are non-decreasing functions with $ \gamma_{r,k}(0)=0 $ and $ \gamma_{r,k}(1)=1 $.  
Note that the signals $ b^*_r(\cdot) $ play the role of common unknown reference shapes, and $w_{r,k}(\cdot)$ are zero mean white Gaussian processes independent for all different~$r$ and $k$.
This perturbation in the coupling model may be understood in two ways: \textit{1)} As some prior knowledge that the coupling relationship between factors $\mB_k$ is not exactly a warping. \textit{2)} As a variable splitting that makes the underlying optimization problem easier to solve. Indeed, if additional constraints are imposed on factors $\mB_k$, for instance nonnegativity, we will show below that the estimation process can be cast as constrained least squares problem.

For discrete time samples $ t_1,\, \cdots,\, t_J $ and assuming $
\gamma_{r,k}(t_j) $ are known, this approach implies that $b_{r,k}(t_j)$  are
independent Gaussian random variables \linebreak $ b_{r,k}(t_j)\sim
\mathcal{N}\left(b^*_r(\gamma_{r,k}(t_j)),\sigma_w^2 \right)  $, where $ \sigma_w^2 $ is a known variance. 
With this prior, criterion~\eqref{mle} can be modified to obtain the following MAP cost function:
\begin{equation}
\mathcal{C}=\mathcal{L} + \frac{1}{\sigma_w^2}\sum_{r,k,j} \Bigl[\mB_{k}(r,j)-b^*_r\bigl(\gamma_{r,k}(t_j)\bigr)\Bigr]^2, \label{map} 
\end{equation}
where the coupling term is introduced by the prior. 
The minimum of $\mathcal{C}$ over all parameters yield the proposed model, coined Registered CP.
The main difference with (\ref{mle}) is that the additional constraints are expected to  solve the rotational ambiguity intrinsic to matrix factorizations.

\noindent It is worth noting that:
\vspace{-6pt}\begin{itemize}
\item \textbf{CP model:} If $\gamma_{r,k}(\cdot)$ are identity and if $\sigma_w^2\rightarrow0$, then the model becomes a CP model obtained by stacking matrices $ \matr{M}_{k} $ along a third dimension.
    
\item \textbf{Indeterminacy:} An indeterminacy remains in determining canonical
    $b^*_r(\cdot) $ and $ \gamma_{r,k}(\cdot) $, since for any given $ r $ one
    can apply a common warping to all $ b_{r,k}(\cdot) $ and obtain a different
    $b^*_r(\cdot) $: $b_{r,k}=\bigl(b_r^*\circ\gamma^{-1}\bigr)\circ \bigl(\gamma \circ\gamma_{r,k} \bigr)$.
In other words, diffeomorphisms $\gamma_{r,k}$ can only be obtained up to a common diffeomorphism.
    
\item \textbf{Linear interpolation:} In theory $ b_{r,k} $, $ b^*_r $ and $ \gamma_{r,k} $ are functions
    of continuous time. In practice we work with discrete time. This means
    exact time transformations $b^{\ast}_{r}(\gamma_{r,k}(t))$ are not actually computed.
    Rather, transformed functions are obtained through linear interpolation.
\end{itemize}

\subsection{Parametric model for the diffeomorphisms}
\label{parametric_warp}
In their non-parametric continuous-time form, the diffeomorphisms $ \gamma_{r,k}(t) $ cannot be handled numerically. 
While it is possible to use dynamic programming to process these diffeomorphisms as non-parametric functions~\cite{srivastava2011registration,rivet2016modeling}, this is typically very sensitive to the noise and time consuming, specially if the dataset is large. 
Therefore, to simplify, we assume that these functions can be modeled with a parametric form, with a small number of parameters. 
Multiple parametrized forms for these functions exist. Here we focus on exponential maps. 

\paragraph{Exponential maps:} 
Since $ \gamma_{r,k}(t) $ are also cumulative distribution functions, they can be defined through their derivatives, which are probability density functions, \textit{i.e.} they are positive and sum to one.
We can define easily such functions by applying the exponential map to any function $ \phi_{r,k}(t) $ defined on $ \left[0,\,1\right] $.
This approach is commonly found in curve registration \cite{ramsay1998curve,james2007curve} and it is also referred as the log-derivative approach \cite{marron2015functional}.
It leads to the following diffeomorphism:
\begin{equation}
\gamma_{r,k}(t)=\left( \int_{0}^{t}e^{\phi_{r,k}(s)}\text{d}s\right) /\left( \int_{0}^{1}e^{\phi_{r,k}(s)}\text{d}s\right).
\label{diff-exp}
\end{equation}

The main purpose of this representation is that we can parametrize the
functions $ \phi_{r,k}(t) $ without imposing monotonicity constraints. In
particular, we can assume that all $ \phi_{r,k}(t) $ are linear
combinations of $ n $ functions $\psi_i(t)$:

\begin{equation}
\phi_{r,k}(t)=\phi(t,\mathbf{\beta}_{r,k})=\sum\limits_{i=1}^{n} \beta_{r,k}^{i} \psi_i(t).
\label{log-derivative}
\end{equation}
where $ \mathbf{\beta}_{r,k}=[\beta_{r,k}^{1}\, \cdots \, \beta_{r,k}^{N}]^{\T}
$ is the vector of parameters characterizing the diffeomorphism. The following
particular cases are of interest:

\textbf{B-splines:} Function $ \psi_i(t) $ can be a B-splines with a fixed number of knots and degree.

\textbf{Linear:} If a simple linear function is used, with $ n=1 $ and $ \psi_1=-t $, then the diffeomorphisms are 
\begin{equation}
\gamma_{r,k}(t)=\dfrac{1-e^{-\beta_{r,k}t}}{1-e^{-\beta_{r,k}}}.
\label{simple_diff}
\end{equation}

\textbf{Constant:} If we use a $ 0 $-th order B-splines basis then we obtain the
    parametrization used implicitly in~\cite{hong2009warped}.

\section{Algorithm}\label{sec:algo}
This section describes the alternating algorithm approach to obtain the
Registered CP model.

\subsection{Multiway array decomposition algorithm}
Given a previous update or guess of $ \vb_r^*(\gamma_{k,r}) $, one can minimize w.r.t. $\mA,\mC$ in an alternating approach using standard linear least squares, while factor $\mB_{k}$ can be retrieved by solving the following least squares problem: 
\begin{equation}\label{eq:penB}
\mB_{k} = \argmin_{\mB=[\vb_1,\dots,\vb_r]} \Bigl\| \mM_k - \mA\mD_k\mB^T \Bigr\|_F^2 
+ \lambda_k \sum\limits_{r=1}^{R}{\Bigl\|\vb_{r,k} - \vb_r^*\bigl[\gamma_{r,k}\bigr] \Bigr\|_F^2 },
\end{equation}
where $\vb^*_r[\gamma_{r,k}]$ stands for $\vb^*_r(\gamma_{r,k}(t))$ taken at sampled times points $t_i$ using linear interpolation

\subsection{Shape Alignment using exponential maps}
From this point onwards, the diffeomorphisms $\gamma_{r,k}$ are assumed to be well modelled as the previously introduced exponential maps $\gamma_{r,k}(t) = (1-e^{-\beta_{r,k} t})/(1-e^{-\beta_{r,k}})$.
Given previous update of latent factors $\vb_{r,k}$, what needs to be estimated are both the values of
$\beta_{r,k}$ and the underlying $\vb_r^*$. Thus, the following
optimization problem needs to be solved for all $r$: 
\begin{equation}\label{opt_align}
   \argmin_{\{\beta_{r,k}\}_k,\vb^*_r} \sum_k {\frac{1}{\sigma_{w}^2} \Bigl\|\vb_{r,k} - \vb^*_{r}[\gamma_{r,k}] \Bigr\|_F^2}.
\end{equation}
Since estimating both the structured mean and $ \gamma_{r,k}(t) $ is cumbersome, as suggested in~\cite{srivastava2011registration}, an alternating strategy is used.
The following can be used independently as a very simple alignment algorithm summarized in Algorithm~\ref{alg:align}: 
\begin{enumerate}
\item Structure mean estimation $\vb_r^*$: Given the values of $\beta_{r,k}$, the structured averages $\vb_r^*$ are computed as the solutions of linear systems, namely
\begin{equation}\label{eq:meanb}
    \vb_r^* = \underset{\vb}{\argmin} \sum_k \bigl\| \vb_{r,k} - \matr{P}_{r,k}\vb \bigr\|_F^2,
\end{equation}
where $\matr{P}_{r,k}$ is the interpolation matrix obtained by linear interpolation from the sampling grid $[t_j]_j$ to the warped sampling grid $[\gamma_{r,k}(t_j)]_j$. 

\item Warping parameters estimation: Given $\vb^*_r$, the criterion~\eqref{opt_align} becomes $K$ one dimensional problems.
And even through it is highly non-convex in the general case, good values of $\beta_{r,k}$ can be computed using a grid search. 
Multiple strategies can then be used to refine the search space once convergence is achieved and we used in particular the Golden Search method~\cite{press2007numerical}. 
In both cases, the cost of one evaluation is rather low since computing~\eqref{opt_align} requires a linear interpolation and $K\times R\times J$ multiplications, but evaluating the cost on a grid can be time consuming. 
\end{enumerate}
This algorithm should converge to a local minimum of the alignment cost function since the cost is reduced at each iteration and for each block of parameters.


\begin{algorithm}
    \begin{algorithmic}
        \STATE \textbf{Input:} Initial target $\vb^*$, initial warping
        parameters $\beta_k$, similar-shaped functions~$\{\vb_k\}_k$, regularization
        parameters $\{\lambda_k\}_k$.
        \WHILE{residual $\sum_k \lambda_k \|\vb_k - \vb^*[\gamma_k] \|_F$ is
        too large}
        \STATE \underline{Structure mean estimation:} set $\vb^*$ as either the 
        \STATE \hspace{6pt} 1. $\vb_{k}[\gamma_k^{-1}]$ that minimized the residuals (first iteration)
        \STATE \hspace{6pt} 2. the solution to~\eqref{eq:meanb}
        \STATE \hspace{6pt} 3. initial target $\vb^*$ (inside a larger optimization scheme)
            \STATE \underline{Warping parameters estimation:} $\forall k$
            \IF{Residuals are higher than some threshold (coarse estimation)}
            \STATE Compute criterion~\eqref{opt_align} on a grid to define an interval $[a_k,b_k]$ surrounding the optimum.
            \ELSE 
            \STATE Find the optimal $\beta_k$ in interval $[a_k,b_k]$ using Golden Search. 
            \ENDIF
        \ENDWHILE
        \STATE \textbf{Output:} Estimated warping parameters $\{\beta_k\}_k$ and
        structured mean $\vb^*$.
    \end{algorithmic}
    \caption{Alignment algorithm under parametrized diffeomorphisms.}\label{alg:align}
\end{algorithm}



\subsection{Detailed 3-way algorithm}
Joining the alternating least squares update of factors $\mA$, $\mB_k$, and $\mC$ with the alignment algorithm (Algorithm 1) leads to Algorithm~\ref{alg:als_RP}, which is given below along with some implementation details.
It can be easily adapted for constrained Registered CP by replacing the least squares solver with a constrained one: e.g., for nonnegative least squares, one can use the algorithm described in \cite{gillis2012accelerated}.

\begin{algorithm}
    \begin{algorithmic}
        \STATE \textbf{Input:} Data matrices $\{\mM_k\}_k$, initial guesses $\mA$,
        $\mC$, $\{\mB_k\}_k$, initial $\{\lambda_k\}_k$ values.
        
        \WHILE{Stopping criterion is not met}
        \STATE $\bullet$ Solve $\argmin_{\mA} \sum_{k}^{K}{ \| \mM_k -
        \mA\mD_k\mB_k^T \|^2_F} $ and normalize column-wise with the $\ell_2$
        norm $\Rightarrow \mA$
        
        \STATE $\bullet$ $\forall k$, solve  $\argmin_{\mD} { \| \mM_k - \mA\mD\mB_k^T \|^2_F},$ $\Rightarrow \{\mD_k\}_k$
        
        \STATE $\bullet$ $\forall k$, solve optimization problem~\eqref{eq:penB} and normalize  column-wise with the $\ell_\infty$ norm, $\Rightarrow\{\mB_k\}_k$
        
        \STATE $\bullet$ Use Algorithm~\ref{alg:align} to align the previously estimated $\{\mB_{k}\}_k$, $\Rightarrow$ $\mB^\ast$ and $\{\beta_{r,k}\}_{r,k}$
        
        \STATE $\bullet$ If necessary, increase the regularization parameters $\Rightarrow \{\lambda_k\}_k$
        \ENDWHILE
        \STATE \textbf{Output:} Estimated factors $\mA$, $\{\mB_k\}_k$ and $\mC$,
        coupling parameters $\mB^*$ and $\{\beta_{r,k}\}_{r,k}$.
    \end{algorithmic}
    \caption{Alternating least squares algorithm for Registered CP under parametrized diffeomorphisms.}\label{alg:als_RP}
\end{algorithm}

\textbf{Initialization: }
 Due to the highly non-convex behavior of the cost function w.r.t. $ \beta_{r,k} $, a good initialization method is required. 
 As a reasonable option, we used the factors given by a standard CP model fitting.
 Moreover, the initial values of $\lambda_k$
are also very important, since large values put too much emphasis on
the regularization terms, which implies factors $B_k$ not change much and the
algorithm mostly fits $\mA$ and $\mC$. Empirically, we used the following
values for the values of $\lambda_k$ at the first and second iterations:
\begin{equation}\label{eq:lambda_init}
    \lambda_k^{0} = 10^{-\frac{SNR}{10}} \frac{\|M_k - A^{0}D^{0}_{k}{B^{0}_{k}}^{T}
    \|_F^2}{\|B^{0}_k \|_F^2} \text{  and  }
    \lambda_k^{1} = 10^{-\frac{SNR}{10}} \frac{\|M_k - A^{1}D^{1}_{k}{B^{1}_{k}}^{T} \|_F^2}{\|B^{1}_k -
    {B^{1}}^{\ast}[\Gamma_k] \|_F^2}
\end{equation}
where $A^{0}$ is the initial value of $A$, $A^{1}$ is the estimate of $A$ after the first
iteration, $B^{\ast}[\Gamma_k]$ is a matrix containing stacked
$b_r^{\ast}[\gamma_{r,k}]$ and SNR refers to the expected Signal to Noise ratio of the whole
tensor data.
We used $ \lambda_k^{1} $ in all following iterations. 

\textbf{Normalization:} Columns of $ \mA $ are normalized with $\ell_2$ norm, while the columns of $ \mB_{k} $ are normalized with $ \ell_\infty $ norm.

\textbf{Case $\gamma_{kr} =\gamma_k$ for all $r$:} It may happen that all components in $\mB_k$ have the
same warping, for instance when the variability generating process affects the data uniformly across the sensors. Such an hypothesis is
actually exploited also in~\cite{hong2009warped} and is an underlying hypothesis of  PARAFAC2. Formally, with parametrized diffeomorphisms, this means that $\beta_{kr} = \beta_{k}$ for all $r$. Then
the alignment algorithm can be slightly modified to improve estimation accuracy
since the number of parameters is reduced. 

\section{Experiments on simulated nonnegative data}\label{sec:expe}
In this section, the Registered CP model is tested on simulated nonnegative
data and compared with similar state-of-the-art models, namely the Shift
PARAFAC model and the PARAFAC2 model. Many data alignment models have been
proposed in the literature, but only those two models align the factors directly
inside the optimization process. 

\textbf{Simulation settings:} After setting the rank $R$, factors $\mA$ and $\mC$ are drawn entry-wise from uniform distributions over $[0,1]$. A latent factor
$\mB^\ast$ is generated column-wise using the exponential map, which mode is
randomly determined but so that all $R$ modes do not overlap. The variances are
also randomly determined. Then, $\beta_{k,r}$ are chosen using affine functions of the $k$
variable with random slope depending on the $r$ variable. Thus each component has its own
warping range. Finally, the $\mB_k$ are generated from $\mB^\ast$ using
exponential maps of parameters $\beta_{k,r}$. Additive Gaussian noise variance is determined from a user-defined SNR using $\sigma_n^k = \sqrt{R}10^{-\frac{\text{SNR}}{20}} $. 

In the following experiment, the total reconstruction error $ \varepsilon_{B} $ on $\mB_k$
\begin{equation}\label{eq:err_b}
    \varepsilon_{B}=\left(\sum_{k=1}^{K}{\| \mB_k -
    \Pi_k\widehat{\mB_k}\|^2_F}\right)/\left(\sum_{k=1}^{K}{\|\mB_k \|_F^2}\right)
\end{equation}
is monitored over $N=50$ experiments. $\Pi_k$ is the best permutation that
matches columns of the estimated $\widehat{\mB_k}$ with the true $\mB_k$. Note
that in~\eqref{eq:err_b}, the $\mB_k$ matrices are normalized column-wise using the
$\ell_2$ norm. The rank is set to $R=3$ and data
dimensions are $15\times 200 \times 10$. The Registered CP algorithm is
initialized by the result of 100 iterations of standard alternating least squares. 

\begin{figure}
    \includegraphics[width=0.8\textwidth]{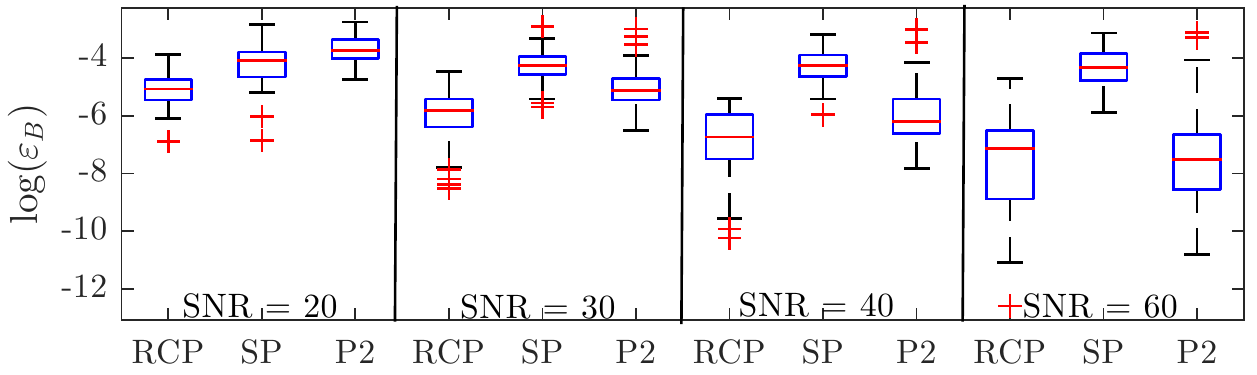}
    \caption{$ \log(\varepsilon_{B}) \; \times \; \text{SNR}$: RCP - registered CP, SP - shift PARAFAC, P2 - PARAFAC2.}\label{fig:err_b_snr}
\end{figure}

Figure~\ref{fig:err_b_snr} shows $ \varepsilon_{B} $ for several SNR values and
the various mentioned algorithms. Although PARAFAC2 algorithm should not
perform well since it relies on the assumption that $\gamma_{k,r}=\gamma_{k}$
for all $r$, it outperforms both the Shift-PARAFAC and the Registered CP
model at high SNR values. However, on average, the Registered CP performs the best for
medium and low SNR values. All algorithm feature a high variability in their
outputs, thus indicating a high sensibility to the initialization. The fact that PARAFAC2 uses the best of several initializations  is probably the reason why it performs best at high SNR.

\begin{figure}
    \includegraphics[width=0.43\textwidth]{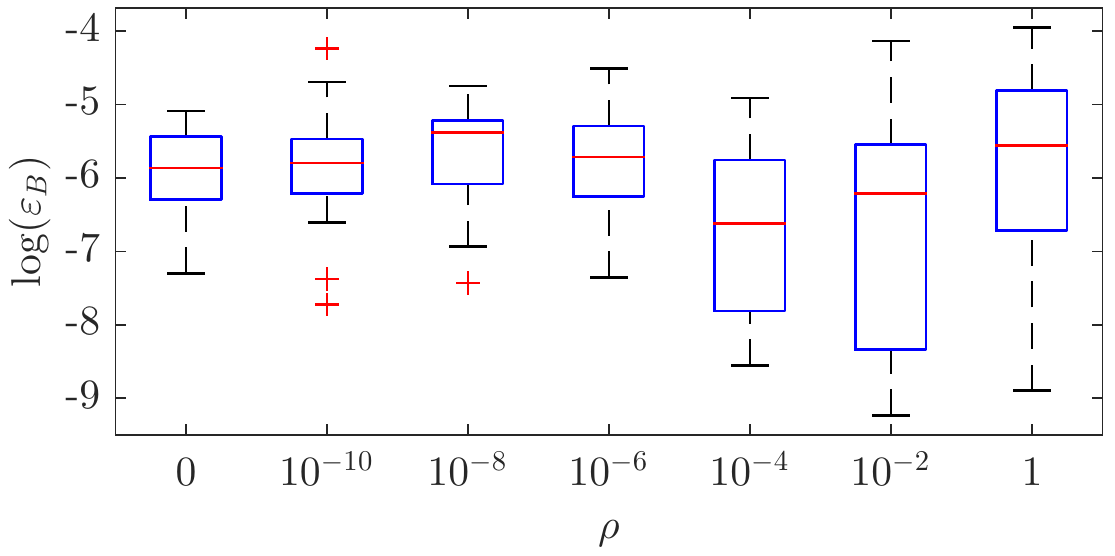}
    \includegraphics[width=0.43\textwidth]{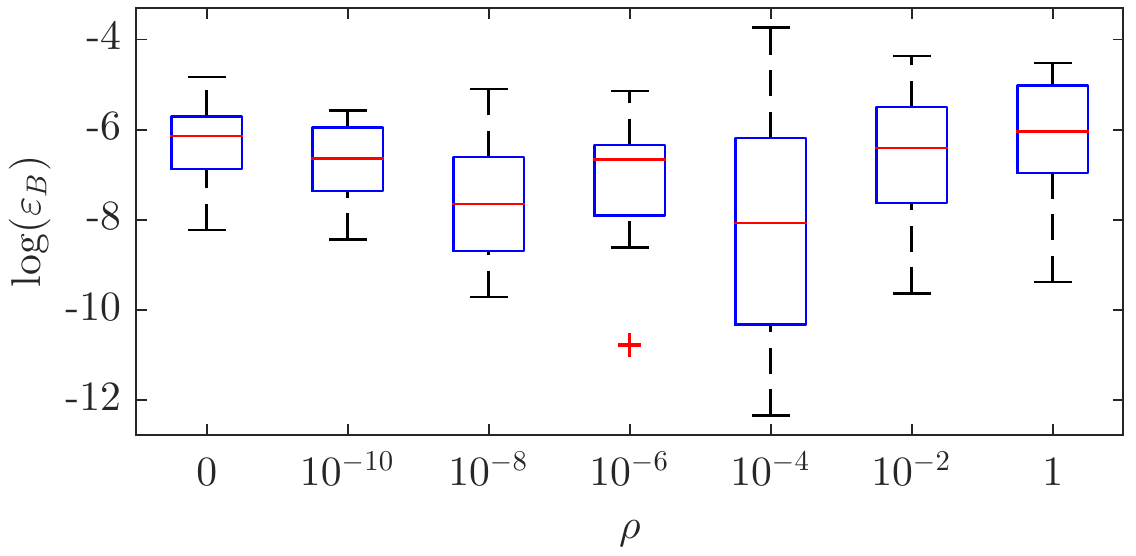}
    \caption{$ \log(\varepsilon_{B}) \; \times \; \rho$: Left - $ \text{SNR}= 40 $,
    right - $ \text{SNR}= 60 $.}\label{fig:err_b_rho}
\end{figure}

In order to study the dependence of $ \varepsilon_{B} $ on regularization parameters $\lambda$ for Registered CP, in a second experiment, instead of the
values suggested in equation~\eqref{eq:lambda_init}, we fixed  values $ \text{SNR}=40 $ or $ \text{SNR}=60 $
gridded over a multiplicative coefficient $\rho$ in front of the initial
$\lambda_k$ values: 
\begin{equation}\label{eq:lambda_init2}
    \lambda_k^{0} = \rho \frac{\|M_k - A^{0}D^{0}_{k}{B^{0}_{k}}^{T}
    \|_F^2}{\|B^{0}_k \|_F^2} \text{  and  }
    \lambda_k^{1} = \rho \frac{\|M_k - A^{1}D^{1}_{k}{B^{1}_{k}}^{T} \|_F^2}{\|B^{1}_k -
    {B^{1}}^{\ast}[\Gamma_k] \|_F^2}.
\end{equation} Figure~\ref{fig:err_b_rho} shows the obtained results for $N=25
$ realizations. It can be observed
that finding a good set of regularization parameters is important to obtain
better results on average. The good performance of the uncoupled matrix factorization algorithm (regularization set to 0) is due to the nonnegativity constraints applied on
all factors. Nevertheless, using the Registered CP model, estimation
performances on the $\mB_k$ are improved at both $ \text{SNR}=40 $ and $ 60 $. Variability however seems to increase alongside the amount of
regularization.

\vspace*{-20pt}

\section{Conclusion}\label{sec:ccl}
A new coupled tensor decomposition model is introduced, namely the Registered
CP model, where factors on one mode are similar up to time
contraction or dilatation. A specific class of diffeomorphisms is used to
generate a decomposition algorithm that can identify both the factors and the
latent coupling parameters. Simulations on synthetic data show encouraging
results, but the Registered CP model is yet to be tested on actual data
sets. Furthermore, in future works, the class of allowed diffeomorphisms should be enlarged.









\bibliographystyle{plain}

\begin{thebibliography}{10}

\bibitem{BeckACGWM15:spmag}
H.~Becker, L.~Albera, P.~Comon, R.~Gribonval, F.~Wendling, and I.~Merlet.
\newblock Brain source imaging: from sparse to tensor models.
\newblock {\em IEEE Sig. Proc. Magazine}, 32(6):100--112, November 2015.

\bibitem{bro1999parafac2}
R.~Bro, C.~A. Andersson, and H.~A.~L. Kiers.
\newblock {PARAFAC2-Part} {II}. modeling chromatographic data with retention
  time shifts.
\newblock {\em J. Chemometr.}, 13(3-4):295--309, 1999.

\bibitem{CabrCC16:tsp}
R.~Cabral~Farias, J.~E. Cohen, and P.~Comon.
\newblock Exploring multimodal data fusion through joint decompositions with
  flexible couplings.
\newblock {\em IEEE Trans. Signal Process.}, 64(18):4830--4844, 2016.

\bibitem{comon2009tensor}
P.~Comon, X.~Luciani, and A.~L.~F. De~Almeida.
\newblock Tensor decompositions, alternating least squares and other tales.
\newblock {\em J. Chemometr.}, 23(7-8):393--405, 2009.

\bibitem{gillis2012accelerated}
N.~Gillis and F.~Glineur.
\newblock Accelerated multiplicative updates and hierarchical als algorithms
  for nonnegative matrix factorization.
\newblock {\em Neural Comput.}, 24(4):1085--1105, 2012.

\bibitem{harshman2003shifted}
R.~A. Harshman, S.~Hong, and M.~E. Lundy.
\newblock Shifted factor analysis—{Part I}: Models and properties.
\newblock {\em J. Chemometr.}, 17(7):363--378, 2003.

\bibitem{harshman1972parafac2}
Richard~A Harshman.
\newblock {PARAFAC2}: Mathematical and technical notes.
\newblock {\em UCLA working papers in phonetics}, 22(3044):122215, 1972.

\bibitem{hong2009warped}
S.~Hong.
\newblock Warped factor analysis.
\newblock {\em J. Chemometr.}, 23(7-8):371--384, 2009.

\bibitem{james2007curve}
G.~M. James.
\newblock Curve alignment by moments.
\newblock {\em Ann. Appl. Stat.}, pages 480--501, 2007.

\bibitem{kneip1992statistical}
A.~Kneip and T.~Gasser.
\newblock Statistical tools to analyze data representing a sample of curves.
\newblock {\em The Annals of Statistics}, pages 1266--1305, 1992.

\bibitem{marini2013scream}
F.~Marini and R.~Bro.
\newblock Scream: A novel method for multi-way regression problems with shifts
  and shape changes in one mode.
\newblock {\em Chemometr. Intell. Lab.}, 129:64--75, 2013.

\bibitem{marron2015functional}
J.~S. Marron, J.~O. Ramsay, L.~M. Sangalli, and A.~Srivastava.
\newblock Functional data analysis of amplitude and phase variation.
\newblock {\em Statistical Science}, 30(4):468--484, 2015.

\bibitem{morup2008shift}
M.~M{\o}rup, L.~K. Hansen, S.~M. Arnfred, L.-H. Lim, and K.~H. Madsen.
\newblock Shift-invariant multilinear decomposition of neuroimaging data.
\newblock {\em NeuroImage}, 42(4):1439--1450, 2008.

\bibitem{press2007numerical}
W.~H. Press, S.~A. Teukolsky, W.~T. Vetterling, and B.~P. Flannery.
\newblock {\em Numerical recipes 3rd edition: The art of scientific computing},
  volume~3.
\newblock Cambridge university press Cambridge, 2007.

\bibitem{ramsay1998curve}
J.~O. Ramsay and X.~Li.
\newblock Curve registration.
\newblock {\em J. R. Stat. Soc. Series B Stat. Methodol.}, 60(2):351--363,
  1998.

\bibitem{rivet2016modeling}
B.~Rivet and J.~E. Cohen.
\newblock Modeling time warping in tensor decomposition.
\newblock In {\em IEEE Sensor Array and Multichannel Signal Processing Workshop
  (SAM 2016)}, pages 1--5. IEEE, 2016.

\bibitem{rivet2015multimodal}
B.~Rivet, M.~Duda, A.~Gu{\'e}rin-Dugu{\'e}, C.~Jutten, and P.~Comon.
\newblock Multimodal approach to estimate the ocular movements during {EEG}
  recordings: a coupled tensor factorization method.
\newblock In {\em Conf. Proc. IEEE Eng. Med. Biol. Soc.}, pages 6983--6986.
  IEEE, 2015.

\bibitem{sakoe1978dynamic}
H.~Sakoe and S.~Chiba.
\newblock Dynamic programming algorithm optimization for spoken word
  recognition.
\newblock {\em {IEEE} Trans. Acoust., Speech, Signal Process.}, 26(1):43--49,
  1978.

\bibitem{seichepine2014soft}
N.~Seichepine, S.~Essid, C.~F{\'e}votte, and O.~Capp{\'e}.
\newblock Soft nonnegative matrix co-factorization.
\newblock {\em IEEE Trans. Signal Process.}, 62(22):5940--5949, 2014.

\bibitem{sidiropoulos2000blind}
N.~D. Sidiropoulos, G.~B. Giannakis, and R.~Bro.
\newblock Blind {PARAFAC} receivers for {DS-CDMA} systems.
\newblock {\em IEEE Trans. Signal Process.}, 48(3):810--823, 2000.

\bibitem{smilde2005multi}
A.~Smilde, R.~Bro, and P.~Geladi.
\newblock {\em Multi-way analysis: applications in the chemical sciences}.
\newblock John Wiley \& Sons, 2005.

\bibitem{srivastava2011registration}
A.~Srivastava, W.~Wu, S.~Kurtek, E.~Klassen, and J.~S. Marron.
\newblock Registration of functional data using {F}isher-{R}ao metric.
\newblock {\em arXiv preprint arXiv:1103.3817}, 2011.

\end{thebibliography}

~\vfill


\end{document}